\documentclass[letterpaper, twocolumn]{IEEEtran}
\ifCLASSINFOpdf
  \usepackage[pdftex]{graphicx}
  \DeclareGraphicsExtensions{.pdf,.jpeg,.png}
\else
   \usepackage[dvips]{graphicx}
\fi
\usepackage[cmex10]{amsmath}
\usepackage{color}
\usepackage{amssymb}
\usepackage{bm}
\usepackage{caption}
\usepackage{lipsum}

\usepackage{float}
\usepackage{placeins}
\usepackage{multirow}
\usepackage{hhline}

\usepackage{algorithm,algorithmic}
\usepackage{array}
\ifCLASSOPTIONcompsoc
  \usepackage[caption=false,font=footnotesize,labelfont=sf,textfont=sf]{subfig}
\else
  \usepackage[caption=false,font=footnotesize]{subfig}
\fi
\usepackage{fixltx2e}

\usepackage{amsthm}
\usepackage [autostyle, english = american]{csquotes}
\MakeOuterQuote{"}
\theoremstyle{plain}

\newcommand{\ds}{\displaystyle}

\newcommand{\la}{\langle}
\newcommand{\ra}{\rangle}

\newcommand{\De}{\Delta}

\newcommand{\mb}{\mathbb}
\newcommand{\mc}{\mathcal}

\newcommand{\tb}{\textbf}
\newcommand{\nn}{\nonumber}
\newcommand{\bea}{\begin{eqnarray}}
\newcommand{\eea}{\end{eqnarray}}
\newcommand{\beq}{\begin{equation}}
\newcommand{\eeq}{\end{equation}}

\newtheorem{ex}{Example}\newcommand{\Ex}{\begin{ex}\rm}
\newcommand{\eex}{\end{ex}}
\begin{document}

\title{Matrix Product State for Feature Extraction of Higher-Order Tensors}
\author{Johann A. Bengua$^1$, Ho N. Phien$^1$, Hoang D. Tuan$^1$\thanks{$^1$Faculty of Engineering and Information Technology,	University of Technology Sydney, Ultimo, NSW 2007, Australia; Email: johann.a.bengua@student.uts.edu.au, ngocphien.ho@uts.edu.au, tuan.hoang@uts.edu.au.} and Minh N. Do$^2$\thanks{
$^2$Department of Electrical and Computer Engineering and the Coordinated Science Laboratory, University of Illinois at Urbana- Champaign, Urbana, IL 61801 USA; Email: minhdo@illinois.edu }}

\maketitle

\vspace*{-1.0cm}

\begin{abstract}
	This paper introduces matrix product state (MPS) decomposition as a computational tool for extracting features of multidimensional data represented by higher-order tensors.
	Regardless of tensor order, MPS extracts its relevant features to the so-called \emph{core tensor} of maximum order three which can be used for classification. Mainly based on a successive sequence of singular value decompositions (SVD), MPS is quite simple to implement without any recursive procedure needed for optimizing local tensors. Thus, it leads to substantial computational savings compared to other tensor feature extraction methods such as higher-order orthogonal iteration (HOOI) underlying the Tucker decomposition (TD). Benchmark results show that MPS can reduce significantly the feature space of data while achieving better classification performance compared to HOOI.
\end{abstract}

\begin{IEEEkeywords}
	Higher-order tensor, tensor feature extraction, supervised learning, tensor classification,
	matrix product state (MPS), core tensor, dimensionality reduction, Tucker decomposition (TD).
\end{IEEEkeywords}

\section{Introduction}\label{sec:introduction}

\IEEEPARstart{T}{here} is an increasing need to handle large multidimensional datasets that cannot efficiently be analyzed or processed using modern day computers.
Due to the curse of dimensionality it is urgent to investigate mathematical tools which can evaluate information beyond the properties of large matrices \cite{Lu20111540}.
The essential goal is to reduce the dimensionality of multidimensional data, represented by tensors, with a minimal information loss by mapping the original
tensor space to a lower-dimensional tensor space through tensor-to-tensor or tensor-to-vector projection \cite{Lu20111540}.
The most natural option for these kinds of projection is to utilize an appropriate tensor decomposition \cite{KOL09} to represent the original tensor
in terms of a combination of possibly lower-order tensors.

A popular method for tensor decomposition is the Tucker decomposition (TD) \cite{TUCKER1966}, also known as higher-order singular value decomposition (HOSVD) when
orthogonality constraints are imposed \cite{LAU00}. As a tensor-to-tensor projection, it is an important tool for solving problems related to feature extraction, feature selection and classification of large-scale
multidimensional datasets in various research fields. Its well-known application in computer vision was introduced in \cite{VA02} to analyze some
ensembles of facial images represented by fifth-order tensors. In data mining, the HOSVD was also applied to identify handwritten digits \cite{SAV07}. In addition, the HOSVD has
been applied in neuroscience, pattern analysis, image classification and signal processing \cite{PHA10,KUA14,CIC15}. The central concept of using the TD is to decompose a large
multidimensional tensor into a set of \emph{common factor} matrices and a single \emph{core tensor} which is considered as reduced features of the original tensor in spite of its lower dimension \cite{PHA10}. In practice, the TD is
often performed in conjunction with some constraints, e.g. nonnegativity, orthogonality, etc., imposed on the common factors in order to obtain a better feature core tensor \cite{PHA10}.
However, constraints like orthogonality often leads to an NP-hard computational problem \cite{LU08}. Practical application of the TD is normally limited to small-order tensors. This is due to the fact that the
TD core tensor preserves the higher-order structure of the original tensor, with its dimensionality remaining fairly large in order to capture relevant interactions between components of the tensor \cite{KOL09}.

The higher-order orthogonal iteration (HOOI) \cite{LAU00_1} is an alternating least squares (ALS)  for
finding TD approximation  of a tensor. Its  application to independent component analysis (ICA) and simultaneous matrix diagonalization was investigated in \cite{DeLathauwer200431}. Another  TD-based method is multilinear principal component analysis (MPCA) \cite{LU08}, an extension of classical principal component analysis (PCA), which is closely related to HOOI. The motivation behind MPCA is that generally PCA takes vectors as inputs for dimensionality reduction, hence tensor data would need to be vectorized and this can result in large computational and memory requirements, even for low order data.

The matrix product state (MPS) decomposition \cite{VER04_1,VID03, VID04,PGA07} is a tensor-to-tensor projection that has been proposed and applied to study quantum many-body systems with great success,
prior to its introduction to the mathematics community under the name tensor-train (TT) decomposition \cite{OSE11}, however, to the best of our knowledge  its application to machine learning and pattern analysis
has not been proposed. The MPS decomposition is fundamentally different from the TD in terms of its geometric structure as it is made up of local component tensors with maximum order three. Consequently, applying the MPS decomposition to large higher-order tensors can potentially avoid the computational bottleneck of the TD and related algorithms.

Motivated by both the TD and MPS decompositions, we propose to use MPS as a dimensionality reduction technique that consists of low-order
common factors and a low-order core tensor. Specifically, MPS decomposes a higher-order tensor in such a way
that its MPS representation is expressed in a \emph{mixed-canonical form} \cite{SCHO2011}. In this form, a unique core tensor can be extracted and is naturally described by an orthogonal space spanned by the common factors. This new approach provides a unique and efficient way of feature extraction applied to tensor classification problem. Specifically, in the tensor classification problem it is applied to firstly extract the core tensor and common factors for the training set. Then the core tensor of test set is extracted by means of common factors. Once the core tensors of both training and test sets are acquired, they can be used for classifiers such as K-nearest neighbors (KNN) and linear discriminant analysis (LDA).

When compared to HOOI, MPS is not only simpler to implement but also more effective in terms of computational savings, feature space and classification success rate (CSR).
This is due to the fact that MPS can obtain orthogonal common factors based on successive SVDs without needing any recursive local optimization procedure. We use supervised learning (classification) problems to benchmark MPS and compare its performance with HOOI. The datasets include the Columbia object image libraries 100 (COIL-100) \cite{NEN96b,PON98},
the brain-computer imagery (BCI) dataset from Shanghai Jiao Tong University \cite{BCM13}, the extended Yale Face Database B (EYFB) from
the Computer Vision Laboratory of the University of California San Diego\cite{GEO01}. Experimental results show that in most cases, MPS provides better CSR compared to HOOI.

The rest of the paper is structured as follows. Section \ref{Problemstatement} introduces mathematical notation and preliminaries used in the paper.
It then formulates the tensor classification problem and describes how to solve it utilizing the TD.
Section \ref{MPSalgorithms} describes in detail how to apply the concept of MPS to tensor classification problem, and subsequently proposes the idea of the MPS core tensor and common factors. MPS is then described in detail, followed by computational complexity analysis. In Section \ref{Ressec}, experimental results are shown to compare MPS to HOOI. The conclusions are given in Section \ref{Conclusions}.

A preliminary result of this work was presented in \cite{Bengua_2015}. In the present paper, we rigorously introduce MPS algorithm with computational complexity analysis. Also, new benchmark results are rigorously compared with those obtained by HOOI to show the advantages of MPS. In this context, we show that MPS can circumvent the problem of unbalanced matricization incurred in HOOI.

\section{Tensor classification}\label{Problemstatement}
To make the paper self-contained we introduce some notations and preliminaries of multilinear algebra \cite{KOL09}. A \textit{tensor} is a multidimensional array and its \textit{order} (also known as way or mode) is the number of dimensions it contains. Zero-order tensors are scalars and denoted by lowercase letters, e.g., $x$. A first-order tensor is a vector and denoted by boldface lowercase letters, e.g., \textbf{x}. A matrix is a second order tensor and denoted by boldface capital letters, e.g., $\textbf{X}$. A higher-order tensor (tensors of order three and above) are denoted by boldface calligraphic letters, e.g., $\bm{\mc{X}}$. Generally, an \textit{N}th-order tensor is denoted as $\bm{\mc{X}}\in\mathbb{R}^{I_1\times I_2\times\cdots\times I_N}$, where each $I_i$ is the dimension of the local subspace $i$. We also denote $x_i$ as the \textit{i}th entry of a vector \tb{x} and $x_{ij}$ as an element of a matrix \tb{X}. Generally, an element of an $N$th-order tensor $\bm{\mc{X}}$ is denoted as $x_{i_1\cdots i_N}$.

A mode-n fiber of a tensor $\bm{\mc{X}}\in\mathbb{R}^{I_1\times I_2\times\cdots\times I_N}$ is defined by fixing all indices but $i_n$ and denoted by \text{\bf x}$_{i_1\ldots i_{n-1}:i_{n+1}\ldots i_{N}}$.

Mode-$n$ matricization (also known as mode-$n$ unfolding or flattening) of a tensor $\bm{\mc{X}}\in\mathbb{R}^{I_1\times I_2\times\cdots\times I_N}$ is the process of unfolding or reshaping the tensor into a matrix $\tb{X}_{(n)}\in\mathbb{R}^{I_n\times (I_1\cdots I_{n-1}I_{n+1}\cdots I_N)}$ by rearranging the mode-$n$ fibers to be the columns of the resulting matrix. Tensor element $(i_1,\ldots, i_{n-1},i_n,i_{n+1},\ldots, i_{N})$ maps to matrix element $(i_n,j)$ such that
\bea
j=1+\sum_{k=1,k\neq n}^{N}(i_k-1)J_k~~\text{with}~~J_k=\prod_{m=1, m\neq n}^{k-1}I_m.
\label{indexj}
\eea

The mode-$n$ product of a tensor $\bm{\mc{X}}\in\mathbb{R}^{I_1\times I_2\times\cdots\times I_N}$ with a matrix $\tb{A}\in\mathbb{R}^{J_n\times I_n}$ results into a new tensor of size $I_1\times\cdots\times I_{n-1}\times J_n\times I_{n+1}\times\cdots\times I_N$ which is denoted as $\mc{X}\times_n A$. Elementwise, it is described by
\bea
(\bm{\mc{X}}\times_n \tb{A})_{i_1\cdots i_{n-1}j_ni_{n+1}\cdots i_N}=\sum_{i_n=1}^{I_n}x_{i_1\cdots i_n\cdots i_N}a_{j_ni_n}.
\eea
The inner product of two tensors $\bm{\mc{X}},\bm{\mc{Y}}\in\mathbb{R}^{I_1\times I_2\times\cdots\times I_N}$ is defined as
\bea
\la\bm{\mc{X}},\bm{\mc{Y}}\ra &=&\sum_{i_1=1}^{I_1}\sum_{i_2=1}^{I_2}\cdots\sum_{i_N=1}^{I_N}x_{i_1i_2\cdots i_{N}}y_{i_1i_2\cdots i_{N}}.
\eea
Accordingly, the Frobenius norm of $\bm{\mc{X}}$ is $||\bm{\mc{X}}||_F = \sqrt{\la\bm{\mc{X}},\bm{\mc{X}}\ra}$.

Having sufficient notations and preliminaries of multilinear algebra, we are considering the following tensor classification problem:

{\bf Tensor classification problem.} {\it Given a set of $K$ training samples represented by \textit{N}th-order tensors $\bm{\mc{X}}^{(k)}\in\mathbb{R}^{I_1\times I_2\times\cdots\times I_N}$ ($k=1,2,\ldots,K$) corresponding to $D$ categories, and a set of $L$ test data $\bm{\mc{Y}}^{(\ell)}\in\mathbb{R}^{I_1\times I_2\times\cdots\times I_N}$ ($\ell=1,2,\ldots,L$), classify the test data into the categories $D$ with high accuracy.}

The problem is usually addressed by the following steps:
\begin{itemize}
	\item \textit{Step 1:} Apply tensor decomposition method to the training set find a set of common factors and corresponding reduced features of each training sample $\bm{\mc{X}}^{(k)}$.
	\item \textit{Step 2:} Extract the reduced features of each test sample $\bm{\mc{Y}}^{(k)}$ in the test set using the common factors in \textit{Step 1}.
	\item \textit{Step 3:} Perform classification based on the reduced features of training and test sets using conventional methods \cite{Duda2000} such as K-nearest neighbors (KNN) and linear discriminant analysis (LDA).
\end{itemize}
For \textit{Step 1}, the authors in \cite{PHA10} proposed methods based on the TD to obtain the common factors and the core tensor from the training set. More specifically, the $K$ training sample tensors are firstly concatenated along the mode $(N+1)$ so that the training set is represented by an $(N+1)$th-order tensor $\bm{\mc{X}}$
defined as
\bea\label{tens1}
\bm{\mc{X}}=[\bm{\mc{X}}^{(1)} \bm{\mc{X}}^{(2)} \cdots \bm{\mc{X}}^{(K)} ]\in\mathbb{R}^{I_1\times I_2\times\cdots\times I_N\times K}.
\eea
TD-based method such as HOOI \cite{LAU00_1} is then applied  to have the approximation
\bea
\bm{\mc{X}}&\approx&\bm{\mc{G}}\times_1\tb{A}^{(1)}\times_2\tb{A}^{(2)}\cdots\times_N\tb{A}^{(N)},
\label{TD}
\eea
where each matrix $\tb{A}^{(j)}= [\tb{a}^{(j)}_{1},\tb{a}^{(j)}_{2},\ldots, \tb{a}^{(j)}_{\Delta_{j}}]\in\mathbb{R}^{I_j\times \Delta_j} (j = 1,2,\ldots, N)$ is orthogonal, i.e. $\tb{A}^{(j)T}\tb{A}^{(j)}=\tb{I}$ ($\tb{I}\in\mathbb{R}^{\Delta_j\times \Delta_j}$ denotes the identity matrix). It is called by a \textit{common factor} matrix and can be thought of as the principal components in each mode $j$. The parameters $\Delta_{j}$ satisfying

\begin{equation}\label{rank1}
\Delta_{j}\leq\text{rank}(\tb{X}_{(j)})
\end{equation}
are referred to as the bond dimensions or compression ranks of the TD. The $(N+1)$th-order tensor
\[
\bm{\mc{G}}\in\mathbb{R}^{\Delta_1\times \Delta_2\times\cdots\times \Delta_N\times K}
\] is called the \textit{core tensor}, which contains reduced features of the training samples, is represented in the subspace spanned by the common factors $\tb{A}^{(j)}$. More specifically, if $\bm{\mc{G}}$ is matricized such that $\tb{G}_{(N+1)}\in\mathbb{R}^{K\times(\Delta_1\Delta_2\cdots\Delta_N)}$, each row $k$ of $\tb{G}$ represents
\bea
N_f=\prod_{j=1}^{N}\Delta_j
\label{NfTD}
\eea
number of reduced features of the corresponding sample $\bm{\mc{X}}^{(k)}$ in the training set.

The core tensor $\bm{\mc{G}}$ and common factors $\tb{A}^{(j)}$ are found as the solution of the following
nonlinear least square
\begin{equation}\label{altopt}
\begin{array}{r}
\ds\min_{\bm{\mc{R}}, \tb{U}^{(j)}}||\bm{\mc{X}}-\bm{\mc{R}}\times_1\tb{U}^{(1)}\times_2\tb{U}^{(2)}\cdots\times_N\tb{U}^{(N)}||_F^2\\
\mbox{subject to}\quad (\tb{U}^{(j)})^{T}\tb{U}^{(j)}=\tb{I}, j=1,...,N,\end{array}
\end{equation}
which is addressed by alternating least square (ALS) in each $\tb{U}^{(j)}$ (with other $\tb{U}^{(\ell)}$, $\ell\neq j$ held fixed).
The computation complexity per one iteration consisting  of $N$  ALS in $\tb{U}^{(j)}$, $j=1,...,N$  is \cite{Mariya11}
\begin{equation}\label{complex}
\mc{O}(K\De I^N + NKI\De^{2(N-1)}+NK\De^{3(N-1)})
\end{equation}
for
\begin{equation}\label{complex1}
I_j\equiv I\quad\mbox{and}\quad \Delta_j\equiv\Delta, j=1, 2,...,N.
\end{equation}

After computing the core tensor and the common factors for the training set, we proceed to \textit{Step 2} to extract the core tensor containing reduced features for the test set. Specifically, the core tensor for the test set is given by
\bea
 \bm{\mc{Q}} = \bm{\mc{Y}}\times_1(\tb{A}^{(1)})^T\cdots\times_N(\tb{A}^{(N)})^T,
\eea
where the test set is defined as
\bea\label{tens2}
\bm{\mc{Y}}=[\bm{\mc{Y}}^{(1)} \bm{\mc{Y}}^{(2)} \cdots \bm{\mc{Y}}^{(L)} ]\in\mathbb{R}^{I_1\times I_2\times\cdots\times I_N\times L},
\eea
and $\bm{\mc{Q}}\in\mathbb{R}^{\Delta_1\times \Delta_2\times\cdots\times \Delta_N\times L}$. Again, the test core tensor $\bm{\mc{Q}}$ can be matricized such that $\tb{Q}_{(N+1)}\in\mathbb{R}^{L\times(\Delta_1\Delta_2\cdots\Delta_N)}$, each row $l$ of $\tb{Q}$ represents $N_f=\prod_{j=1}^{N}\Delta_j$ number of reduced features of the corresponding sample $\bm{\mc{Y}}^{(l)}$ in the test set. Finally, $\tb{G}_{(N+1)}$ and $\tb{Q}_{(N+1)}$ can be used for performing classification according to \emph{Step 3}.

Although the core tensors $\bm{\mc{G}}$ and $\bm{\mc{Q}}$ can be used for direct training and classification in \emph{Step 3}, their dimensionality often remain large. This is due to the fact that they retain the same orders of their own original tensors. Thus, it may require a further dimensionality reduction of the core tensors using techniques such as Fisher score ranking before inputting them into classifiers to improve the classification accuracy\cite{PHA10}. Besides, the computational complexity of
HOOI to obtain (\ref{TD}) is high as (\ref{complex}) for the computational  complexity per one iteration shows,
which may become prohibitive when the order of the training tensor is large. In addition, due to the unbalanced single-mode matricization (one mode versus the rest) of the tensor when using HOOI, it may not be capable of capturing the mutual correlation between modes of the tensor which usually needs multimode matricization (a few modes versus the rest). Hence, it might cause loss of important information for classification while decomposing the tensor. To circumvent these issues, we propose to use the MPS decomposition in the next section.

\section{Matrix Product State Decomposition for Tensor Feature Extraction}\label{MPSalgorithms}
In this section we develop a tensor feature extraction  method based on MPS decomposition as an alternative solution to the above stated tensor classification problem.
 Subsection \ref{MPS1} presents the concept of common factors and a core tensor for the MPS decomposition of tensor feature extraction and classification problems. Then the MPS is proposed in Subsection \ref{MPS2}.
We finally analyse the computational complexity of MPS in comparison with HOOI in Subsection \ref{Complexity}.

\subsection{Common factors and core tensor of matrix product state decomposition}\label{MPS1}
We now introduce the concept of core tensor and common factors of an MPS representing the training tensor  $\bm{\mc{X}}$ in (\ref{tens1}). Without loss of generality, let us opt to \emph{permute} the mode $K$ of the tensor $\bm{\mc{X}}$
such that
\[
\bm{\mc{X}}\in\mathbb{R}^{I_1\times\cdots I_{n-1}\times K\times I_{n}\cdots\times I_{N}}
\]
(the mode $K$ is now located at $n$ which can be chosen arbitrarily but conventionally we choose it at the middle of the chain, say $n=\text{round}(N/2)$). Accordingly, positions of modes $I_n,\ldots, I_N$ are shifted by one to the right. Then, we present the elements of $\bm{\mc{X}}$ in the following \emph{mixed-canonical form} \cite{SCHO2011} of the matrix product state (MPS) or tensor train (TT) decomposition \cite{PGA07,VID03, VID04,OSE11}:
\beq
x_{i_1\cdots k\cdots i_{N}}=\tb{B}^{(1)}_{i_1}\cdots\tb{B}^{(n-1)}_{i_{n-1}}\tb{G}^{(n)}_{k}\tb{C}^{(n+1)}_{i_{n}}\cdots\tb{C}^{(N+1)}_{i_{N}},
\label{mpsdef_mixed}
\eeq
where $\tb{B}^{(j)}_{i_{j}}$ and $\tb{C}^{(j)}_{i_{(j-1)}}$ (the upper index "$(j)$" denotes the position $j$ of the matrix in the chain) of dimension $\Delta_{(j-1)}\times\Delta_j$ ($\Delta_{0}= \Delta_{N+1}=1$),
are called "left" and "right" \textit{common factors} which satisfy the following orthogonality conditions:
\bea
\label{LeftCan1}
\sum_{i_j}(\tb{B}^{(j)}_{i_j})^{T}\tb{B}^{(j)}_{i_j}&=&\tb{I},\quad(j=1,\ldots,n-1)	
\eea
and
\bea
\label{RightCan1}
\sum_{i_{(j-1)}}\tb{C}^{(j)}_{i_{(j-1)}}(\tb{C}^{(j)}_{i_{(j-1)}})^{T}&=&\tb{I},~~(j=n+1,\ldots,N+1)~~~
\eea
respectively, where $\tb{I}$ denotes the identity matrix. Each $\tb{G}^{(n)}_{k}$ for $k=1, 2,...,K$ is a matrix of dimension $\Delta_{n-1}\times\Delta_n$ and the MPS \textit{core tensor} is defined by
\[\bm{\mc{G}}^{(n)}=[\tb{G}^{(n)}_{1}\ \tb{G}^{(n)}_{2}\cdots\tb{G}^{(n)}_{K}]\in\mathbb{R}^{\De_{n-1}\times\De_n\times K}
 \]
which describes the reduced features of the training set. The parameters $\Delta_{j}$ are called the bond dimensions or compression ranks of the MPS.

Using the common factors $\tb{B}^{(j)}_{i_{j}}$ and $\tb{C}^{(j)}_{i_{(j-1)}}$, we can extract the core tensor for the test tensor $\bm{\mc{Y}}$. Specifically, we first need to permute $\bm{\mc{Y}}$ defined in Eq.~(\ref{tens2}) in such a way that the index $\ell$ is at the same position as $k$ in the training tensor to ensure the compatibility between the training and test tensors, i.e., the permuted \[
\bm{\mc{Y}}\in\mathbb{R}^{I_1\times \cdots I_{n-1}\times L \times I_{n}\cdots\times I_{N}}.
\]
Then the core tensor $\bm{\mc{Q}}^{(n)}$ of the test tensor $\bm{\mc{Y}}$
 is given by
 \[\bm{\mc{Q}}^{(n)}=[\tb{Q}^{(n)}_{1}\ \tb{Q}^{(n)}_{2}\cdots\tb{Q}^{(n)}_{L}]\in\mathbb{R}^{\De_{n-1}\times\De_n\times L},
 \]
where
\bea
\label{testcoreExtract}
Q^{(n)}_{(l)}&=&\sum_{i_1,\ldots, i_{N}}(\tb{B}^{(1)}_{i_1})^{T}\cdots (\tb{B}^{(n-1)}_{i_{n-1}})^{T}y_{i_1\cdots l\cdots i_{N}}\nn\\&&(\tb{C}^{(n+1)}_{i_{n}})^{T}\cdots (\tb{C}^{(N+1)}_{i_{N}})^{T}.
\eea
Note that as the core tensors $\bm{\mc{G}}^{(n)}$ and $\bm{\mc{Q}}^{(n)}$ of both training and test tensors are extracted by using the same common factors,
they are represented by the same base. Thus, they can be used for the classification directly for \textit{Step 3}. More precisely, we can matricize $\bm{\mc{G}}^{(n)}$ and $\bm{\mc{Q}}^{(n)}$ to
\[
\bm{G}^{(n)}\in\mathbb{R}^{K\times(\De_{n-1}\De_{n})}
 \]
and
\[
\bm{Q}^{(n)}\in\mathbb{R}^{L\times(\De_{n-1}\De_{n})}
\]
such that each of their rows, either
$\bm{G}^{(n)}$ or $\bm{Q}^{(n)}$ is a sample containing
\bea
N_f=\De_{n-1}\De_n
\label{NfMPS}
\eea
number of reduced features of the original sample in either training or test set, respectively.

In the next section we will show that Eq.~(\ref{mpsdef_mixed}) can be implemented straightforwardly without any recursive local optimization procedure like ALS in HOOI required. This results into substantial computational savings. Thus, the MPS can overcome the aforementioned issues of HOOI.

\subsection{Matrix product state for feature extraction}\label{MPS2}	
We describe the MPS method for computing the core tensor and common factors of the training set. More specifically, we show how to decompose the training tensor $\bm{\mc{X}}$ into the MPS according to Eq.~(\ref{mpsdef_mixed}). To this end, we apply two successive sequences of SVDs to the tensor which include left-to-right sweep for computing the left common factors $\tb{B}^{(1)}_{i_1},\ldots,\tb{B}^{(n-1)}_{i_{n-1}}$
and right-to-left sweep for computing the right common factors $\tb{C}^{(n+1)}_{i_{n}},\ldots,\tb{C}^{(N+1)}_{i_{N}}$ and core tensor $\bm{\mc{G}}^{(n)}$ in Eq.~(\ref{mpsdef_mixed})
explained in the following\cite{SCHO2011}:

\textbullet~~\emph{Left-to-right sweep for left factor computation:}	
	
The left-to-right sweep involves acquiring matrices $\tb{B}^{(j)}_{i_j}$ ($i_j=1,\ldots,I_j,~\text{where}~j = 1,\ldots,n-1$) fulfilling orthogonality condition in Eq. (\ref{LeftCan1}). Let us start by performing the mode-1 matricization of $\bm{\mc{X}}$ to obtain the matrix \[
\tb{W}\in\mb{R}^{I_1\times(I_2\cdots K\cdots I_{N})}.
\]
Then applying the SVD to $\tb{W}$ such that
\[\tb{W}= \tb{U}\tb{S}\tb{V}^{T}.
\]
We then define the first common factors $\tb{B}^{(1)}_{i_1} = \tb{U}_{i_1}\in\mb{R}^{1\times\De_1}$, where $\De_1\leq\text{rank}(\tb{X}_{(1)})$, satisfying the left-canonical constraint in Eq. (\ref{LeftCan1}) due to the SVD. In order to find the next common factors $\tb{B}^{(2)}_{i_2}$ we firstly form the matrix
\[
\tb{W} = \tb{S}\tb{V}^{T}\in\mb{R}^{\De_1\times(I_2\cdots K\cdots I_{N})}.
\]
The matrix $\tb{W}$ is then reshaped to
\[
\tb{W}\in\mb{R}^{(\De_1I_2)\times(I_3\cdots K\cdots I_{N})}
 \]
and its SVD is given by
\[
\tb{W}= \tb{U}\tb{S}\tb{V}^{T}.
\]
Reshape the matrix
\[
\tb{U}\in\mb{R}^{(\De_1I_2)\times\De_2}\quad
 (\De_2\leq\text{rank}(\tb{W}))
\]
into a third-order tensor
\[
\bm{\mc{U}}\in\mb{R}^{\De_1\times I_2\times\De_2}
\]
and we define $\tb{B}^{(2)}_{i_2} = \tb{U}_{i_2}$ satisfying the left-canonical constraint due to the SVD. Applying a same procedure for determining $\tb{B}^{(3)}_{i_3}$ by forming a new matrix
\[
\tb{W} = \tb{S}\tb{V}^{T}\in\mb{R}^{\De_2\times(I_3\cdots K\cdots I_{N})},
\]
reshaping it to
\[
\tb{W}\in\mb{R}^{(\De_2I_3)\times(I_4\cdots K\cdots I_{N})},
\]
performing SVD and so on. This procedure is iterated until we obtain all the matrices $\tb{B}^{(j)}_{i_j}$ ($i_j=1,\ldots,I_j,~\text{where}~j = 1,\ldots,n-1$) fulfilling the left-canonical constraint in Eq. (\ref{LeftCan1}).

In a nutshell, after completing the left-to-right sweep elements of tensor $\bm{\mc{X}}$ are written in the following form:
\bea
x_{i_1\cdots i_{n-1}ki_n\cdots i_{N+1}}=\tb{B}^{(1)}_{i_1}\cdots \tb{B}^{(n-1)}_{i_{n-1}} \tb{W}_{(ki_n\cdots i_{N})},
\label{l7}
\eea
where the matrix $\tb{W}$ is reshaped to the matrix form $\tb{W}\in\mb{R}^{(\De_{n-1}K\cdots I_{N-1})\times I_{N}}$ for the next right-to-left sweeping process.

\textbullet~~\emph{Right-to-left sweep for right factor computation:}

Similar to left-to-right sweep, we perform a sequence of SVDs starting from the right to the left of the MPS to get the  matrices $\tb{C}^{(j)}_{i_{(j-1)}}$ ($i_{(j-1)}=1,\ldots,I_{(j-1)},~\text{where}~j = n+1,\ldots,N+1$) fulfilling the right-canonical condition in Eq. (\ref{RightCan1}). To start, we apply the SVD to the matrix $\tb{W}$ obtained previously in the left-to-right sweep such that
\[
\tb{W}= \tb{U}\tb{S}\tb{V}^{T}.
\]
Let us then define
\[\tb{C}^{(N+1)}_{i_{N}} = \tb{V}^{T}_{i_{N}}\in\mb{R}^{\De_N\times 1},
\]
where $\De_N \leq \text{rank}(\tb{W})$, which satisfies the right-canonical constraint (Eq. (\ref{RightCan1})) due to the SVD.  Next, multiply $\tb{U}$ and $\tb{S}$ together and reshape the resulting matrix into
\[
\tb{W}\in\mb{R}^{(\De_{n-1}K\cdots I_{N-2})\times (I_{N-1}\De_N)}.
\]
Again, applying the SVD to the matrix $\tb{W}$, we have
\[
\tb{W}= \tb{U}\tb{S}\tb{V}^{T}.
\]
Reshape the matrix
\[
\tb{V}^{T}\in\mb{R}^{\De_{N-1}\times(I_{N-1}\De_N)},
\]
where $\De_{N-1} \leq \text{rank}(\tb{W})$, into a third-order tensor
\[
\bm{\mc{V}}\in\mb{R}^{\De_{N-1}\times I_{N-1}\times\De_N}
\]
and we define the next common factor $\tb{C}^{(N)}_{i_{(N-1)}} = \tb{V}_{i_{(N-1)}}$ satisfying Eq. (\ref{RightCan1}). We needs to obtain the matrix $\tb{W}$ by multiplying $\tb{U}$ and $\tb{S}$ together for determining the next common factor, i.e. $\tb{C}^{(N-1)}_{i_{N-2}}$. This procedure is iterated until all the common factors $\tb{C}^{(j)}_{i_{(j-1)}}$ ($i_{(j-1)}=1,\ldots,I_{(j-1)},~\text{where}~j = n+1,\ldots,N+1$) are acquired. In the end, we obtain Eq.~(\ref{mpsdef_mixed}) for MPS decomposition of the tensor $\bm{\mc{X}}$ where the core tensor
\[
\bm{\mc{G}}^{(n)}\in\mathbb{R}^{\De_{n-1}\times\De_n\times K}
\]
is determined by reshaping the matrix
\[
\tb{G}^{(n)} = \tb{U}\tb{S}\in\mathbb{R}^{\De_{n-1}\times(K\De_n)}.
\]
Having done this, we can substitute the common factors into Eq.~(\ref{testcoreExtract}) to extract the core tensor for the test tensor.

Note that the MPS decomposition described by Eq. (\ref{mpsdef_mixed}) can be performed exactly or approximately depending on the bond dimensions $\Delta_{j}$ $(j=1,\ldots, N)$ which have the following bound\cite{OSE11}:
\begin{eqnarray}
\Delta_{j}\leq\text{rank}(\tb{W})&\leq&\text{rank}(\tb{X}_{[j]})\label{bound},\label{bound2}
\end{eqnarray}
versus their counterpart (\ref{rank1}) in HOOI, where $\text{rank}(\tb{X}_{[j]})$ denotes the rank of the matrix $\tb{X}_{[j]}$ of size $(I_1I_2\cdots I_j)\times(I_{j+1}\cdots K \cdots I_{N})$ which is the \emph{mode-$(1,2,\ldots, j)$ matricization} of the tensor $\mc{X}$. In practice, each bond dimension $\Delta_{j}$ is usually truncated to be smaller than $\text{rank}(\tb{W})$ on every SVD of $\tb{W}$ leading to an efficient MPS decomposition. To this end, we rely on thresholding the singular values of $\tb{W}$. For instance, applying SVD to the matrix $\tb{W}\in\mathbb{R}^{I\times J}$ (let us assume $I\leq J$), we have $\tb{W}= \tb{U}\tb{S}\tb{V}^{T}$, where $ \tb{S}= diag(s_1,s_2,\ldots,s_I)$ are the nonvanishing  singular values. With a threshold $\epsilon$ being defined in advance, we truncate the bond dimension by keeping only $\De$ singular values such that
\bea
\frac{\sum_{j=1}^{\De}s_j}{\sum_{j=1}^{I}s_j}\geq\epsilon.
\label{BondThreshold}
\eea
Having done this, we have $\tb{W}\approx \tilde{\tb{U}}\tilde{\tb{S}}\tilde{\tb{V}}^{T}$, where $\tilde{\tb{U}}\in\mathbb{R}^{I\times\De}$, $\tilde{\tb{S}}\in\mathbb{R}^{\De\times\De}$ and $\tilde{\tb{V}}^{T}\in\mathbb{R}^{\De\times J}$. Note that the larger the $\epsilon$ the more accurate MPS decomposition of the tensor $\bm{\mc{X}}$ but less efficient in reducing the dimensionality of the tensor. Therefore, one needs to choose an appropriate value for $\epsilon$ via empirical simulations. A summary of applying MPS decomposition for tensor feature extraction can be found in Table \ref{Train_MPS}.
\begin{table}[!thb]
	\centering
	\caption{Matrix product state for tensor feature extraction}
	\label{Train_MPS}	
	\begin{tabular}{*2l} 
		\hline
		\tb{Input:}& $\bm{\mc{X}}\in\mathbb{R}^{I_1\times \cdots \times I_{n-1}\times K\cdots\times I_{N}}$,\\
		&$\epsilon$: SVD threshold\\
		\tb{Output:}& $\bm{\mc{G}}^{(n)}\in\mathbb{R}^{\De_{n-1}\times\De_{n}\times K}$,\\
				   ~& $\tb{B}^{(j)}_{i_j}$ ($i_j=1,\ldots,I_j, j = 1,\ldots,n-1$)\\
				   ~& $\tb{C}^{(j)}_{i_{(j-1)}}$ ($i_{(j-1)}=1,\ldots,I_{(j-1)}, j = n+1,\ldots,N+1$)\\
		\hline		
		\multicolumn{2}{l}{1:~~Set $\tb{W}=\tb{X}_{(1)}$~~~~~~~~~~~~~$\%$ Mode-1 matricization of $\bm{\mc{X}}$}\\
		\multicolumn{2}{l}{2:~~\tb{for} $j=1$ \tb{to} $n-1$ ~~~~$\%$ Left-to-right sweep}\\		
		\multicolumn{2}{l}{3:~~~~~~$\tb{W}= \tb{U}\tb{S}\tb{V}^{T}$~~~~~~~~~~$\%$ SVD of $\tb{W}$}\\
		\multicolumn{2}{l}{4:~~~~~~$\tb{W}\approx \tilde{\tb{U}}\tilde{\tb{S}}\tilde{\tb{V}}^{T}$~~~~~~~~~~$\%$ Thresholding $\tb{S}$ using Eq.~(\ref{BondThreshold})}\\
		\multicolumn{2}{l}{5:~~~~~~$\tb{B}^{(j)}_{i_j} = \tilde{\tb{U}}_{i_j}$~~~~~~~~~~~~$\%$ Set common factors}\\
		\multicolumn{2}{l}{6:~~~~~~$\tb{W}=\tilde{\tb{S}}\tilde{\tb{V}}^{T}$~~~~~~~~~~~~~$\%$ Construct new matrix $\tb{W}$}\\		
		\multicolumn{2}{l}{7:~~\tb{end}}\\
		\multicolumn{2}{l}{8:~~Reshape $\tb{W}\in\mb{R}^{(\De_{n-1}K\cdots I_{N})\times I_{N}}$}\\
		\multicolumn{2}{l}{9:~~\tb{for} $j=N+1$ \tb{down to} $n+1$ ~~$\%$ right-to-left sweep}\\		
		\multicolumn{2}{l}{10:~~~~~$\tb{W}= \tb{U}\tb{S}\tb{V}^{T}$~~~~~~~~~$\%$ SVD of $\tb{W}$}\\
		\multicolumn{2}{l}{11:~~~~~$\tb{W}\approx \tilde{\tb{U}}\tilde{\tb{S}}\tilde{\tb{V}}^{T}$~~~~~~~~~$\%$ Thresholding $\tb{S}$ using Eq.~(\ref{BondThreshold})}\\
		\multicolumn{2}{l}{12:~~~~~$\tb{C}^{(j)}_{i_{(j-1)}} = \tilde{\tb{V}}^{T}_{i_{(j-1)}}$~~~~~~~~~~$\%$ Set common factors}\\
		\multicolumn{2}{l}{13:~~~~~$\tb{W}=\tilde{\tb{U}}\tilde{\tb{S}}$~~~~~~~~~~~~~~$\%$ Construct new matrix $\tb{W}$}\\		
		\multicolumn{2}{l}{14:~\tb{end}}\\
		\multicolumn{2}{l}{15:~Set $\bm{\mc{G}}^{(n)} = \bm{\mc{W}}$~~~~~~~~$\%$ Training core tensor}\\	
		\hline
		\multicolumn{2}{l}{Texts after symbol "$\%$" are comments.}		
	\end{tabular}
\end{table}

\subsection{Complexity analysis}\label{Complexity}
For a given training tensor $\bm{\mc{X}}\in\mathbb{R}^{I_1\times I_2\times\cdots\times I_N\times K}$ under the assumption (\ref{complex1}), the TD and MPS representation of $\bm{\mc{X}}$ consists of
\[
NI\Delta+K\Delta^N
\]
and
\[
(N-2)I\Delta^2+K\Delta^2 +2I\Delta
\]
parameters, respectively. The dominant computational complexity of MPS is $\mc{O}(KI^{(N+1)})$ due to the first SVD of the matrix obtained from mode-1 matricization of $\bm{\mc{X}}$. On the other hand, the computational complexity of HOOI is
(\ref{complex}) per iterations with unknown iteration number to attained the convergence of ALS rounds.
 In addition, it usually employs the HOSVD to initialize the tensors which involves the cost of order $\mc{O}(NKI^{N+1})$, and thus very expensive with large $N$ compared to MPS.

\section{Experimental results}\label{Ressec}
In this section, we apply MPS to perform feature extraction for classification problem. The method is applied to a few datasets and compared with one of the most popular feature extraction methods, i.e. HOOI. Specifically, three datasets, namely Columbia Object Image Libraries (COIL)-100 \cite{NEN96b,PON98}, Extended Yale Face Database B (EYFB) \cite{GEO01} and BCI Jiaotong dataset (BCIJ) \cite{BCM13}, are used to benchmark the simulation. In all simulations, we rely on the threshold $\epsilon$ defined in Eq.~(\ref{BondThreshold}) to adjust the bond dimensions of MPS in Eq.~(\ref{mpsdef_mixed}) as well as that of HOOI in Eq.~(\ref{TD}).

\begin{figure}[htpb!]
	\centering
	\includegraphics[width =\columnwidth]{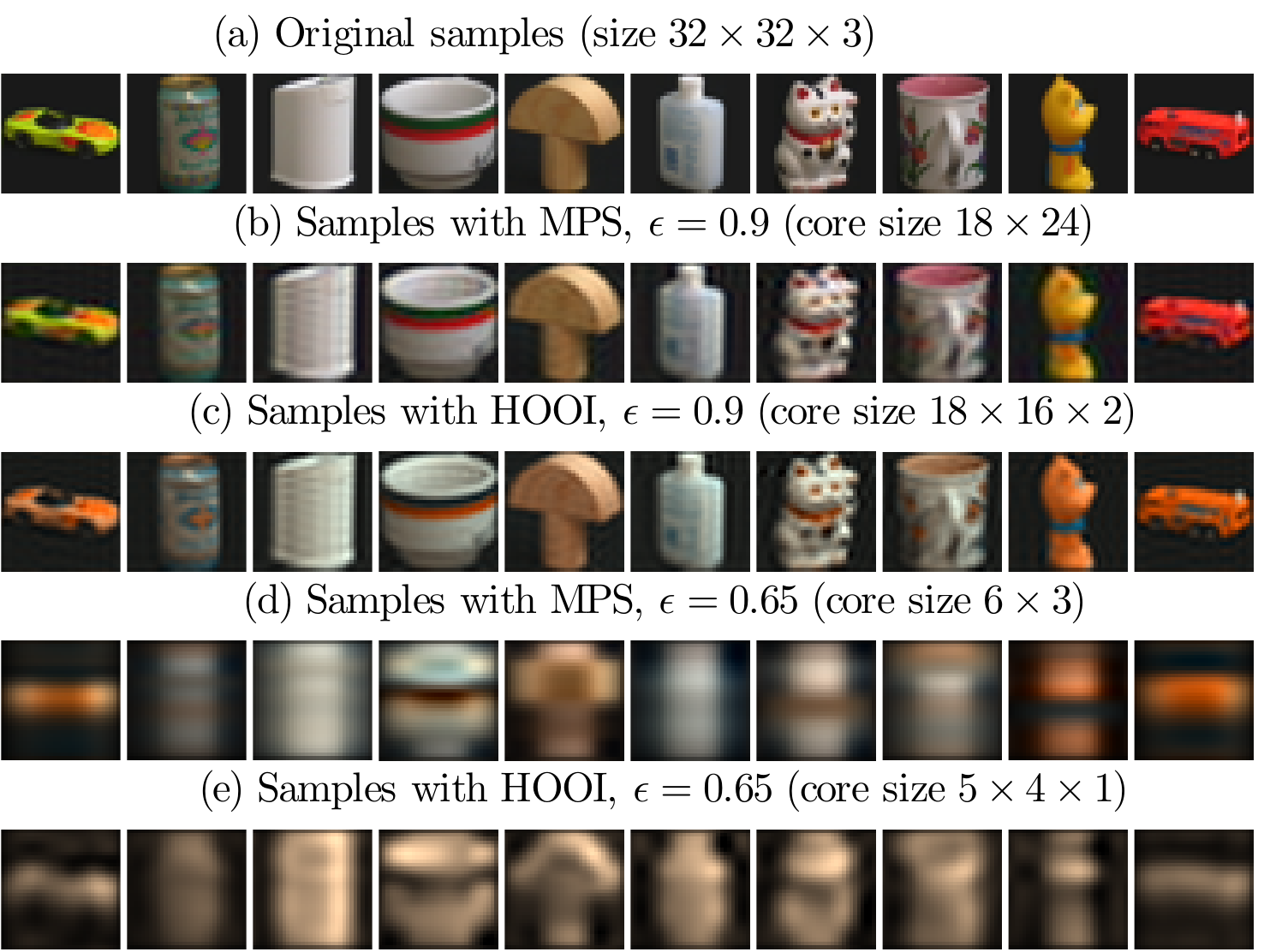}\\
	\caption{Modification of a 10 objects in the {\bf training set} of COIL-100 are shown after applying MPS and HOOI corresponding to $\epsilon=0.9$ and $0.65$ to reduce the number of features of each object.}
	\label{fig1}
\end{figure}

\begin{figure}[htpb!]
	\centering
	\includegraphics[width =\columnwidth]{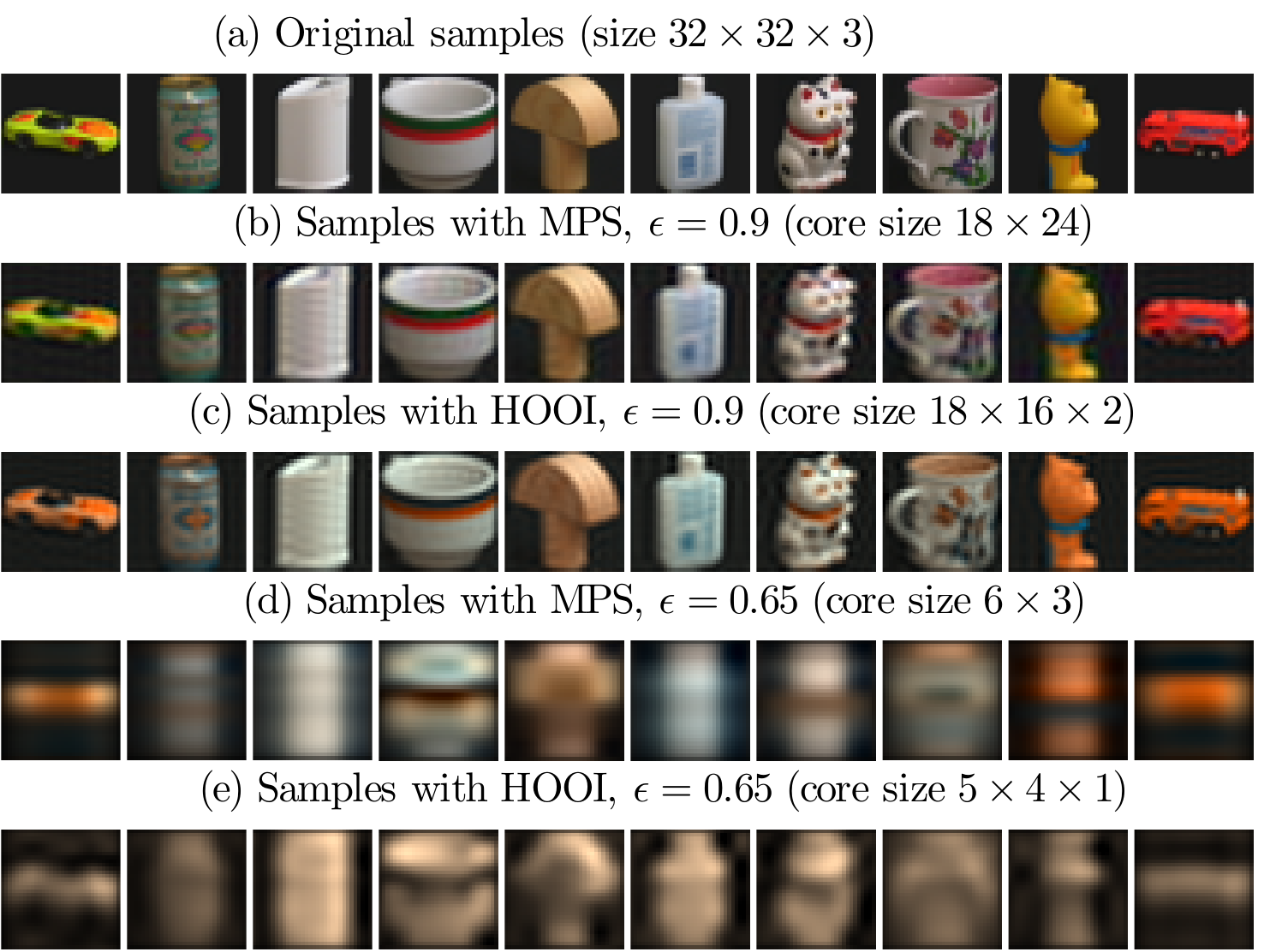}\\
	\caption{Modification of 10 objects in the {\bf test set} of COIL-100 are shown after applying MPS and HOOI corresponding to $\epsilon=0.9$ and $0.65$ to reduce the number of features of each object.}
	\label{fig2}
\end{figure}

The COIL-100 dataset has 7200 color images of 100 objects (72 images per object) with different
reflectance and complex geometric characteristics.
Each image is initially a 3rd-order tensor of dimension $128\times128\times3$ and then is downsampled to the one of dimension $32\times32\times3$. The dataset is divided into training and test sets randomly consisting of $K$ and $L$ images, respectively according to a certain holdout (H/O) ratio $r$, i.e. $r=\frac{L}{K}$. Hence, the training and test sets are represented by four-order tensors of dimensions $32\times32\times3\times K$ and $32\times32\times3 \times L$, respectively. In Fig.~\ref{fig1} and Fig.~\ref{fig2}, we show how a few objects of the training and test sets ($r=0.5$ is chosen), respectively, change after applying MPS and HOOI to reduce the number of features with two different values of threshold, $\epsilon=0.9,0.65$. In both training and test sets, we can see that with $\epsilon=0.9$, the images are not modified significantly due to the fact that many features are preserved. However, in the case that $\epsilon=0.65$, the images are blurred. That is because less features are kept. However, we can observe that the shapes of objects are still preserved. Especially, in most cases MPS seems to preserve the color of the images better than HOOI. This is because the bond dimension corresponding to the color mode $I_3=3$ has a small value, e.g. $\De_3=1$ for $\epsilon=0.65$ in HOOI. This problem arises due to the  the unbalanced matricization of the tensor corresponding to the color mode. Specifically, if we take a mode-3 matricization of tensor $\bm{\mc{X}}\in\mb{R}^{32\times32\times3\times K}$, the resulting matrix of size $3\times(1024K)$ is extremely unbalanced. Therefore, when taking SVD with some small threshold $\epsilon$, the information corresponding to this color mode may be lost due to dimension reduction. On the contrary, we can efficiently avoid this problem in MPS by permuting the tensor such that $\bm{\mc{X}}\in\mb{R}^{32\times K\times3\times32}$ before applying the tensor decomposition.

\begin{figure}[htpb!]
	\centering
	\includegraphics[width =\columnwidth]{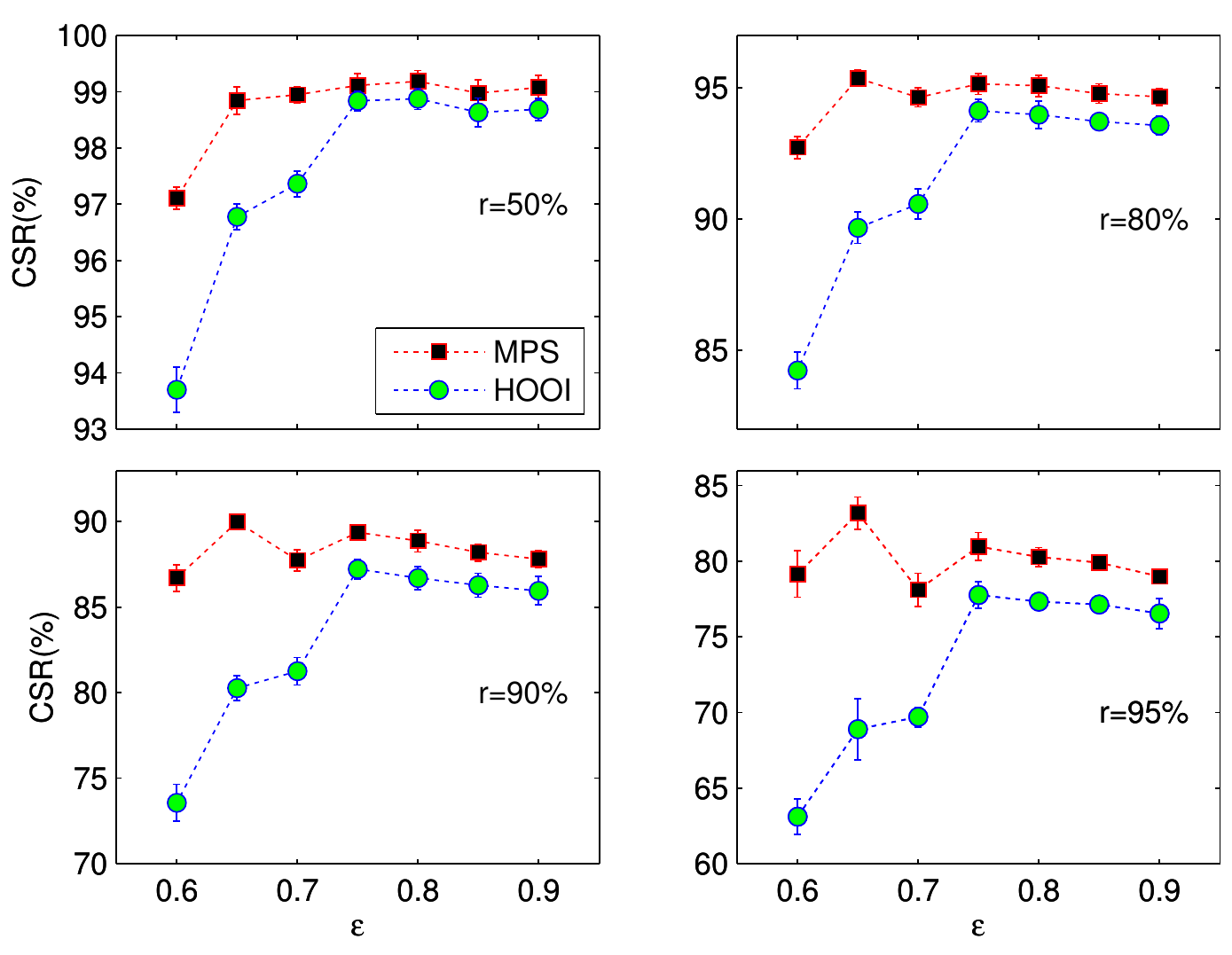}\\
	\caption{Error bar plots of CSR versus thresholding rate $\epsilon$ for different H/O ratios.}
	\label{fig3}
\end{figure}

To validate MPS for classification, the core tensors with full sizes obtained from both methods are input directly to a classifier which is chosen as the K nearest neighbor with K=1 (KNN-1) in our case. For each H/O ratio, the classification success rate (CSR) is averaged over 10 iterations of randomly splitting the dataset into training and test sets. Comparison of performance between MPS and HOOI is shown in Fig.~\ref{fig3} for four different H/O ratios, i.e. $r=(50\%,80\%,90\%,95\%)$. In each plot, we show the CSR with respect to threshold $\epsilon$. We can see that MPS performs quite well when compared to HOOI. Especially, with small $\epsilon$, MPS performs much better than HOOI. Besides, we also show the best CSR corresponding to each H/O ratio obtained by different methods in Table.~\ref{tableCOIL100}. It can be seen that MPS always gives better results than HOOI even in the case of small value of $\epsilon$ and number of features $N_f$
defined by (\ref{NfTD}) and (\ref{NfMPS}) for HOOI and MPS, respectively.

\begin{table*}[htpb!]
	\caption{COIL-100 benchmark: The best CSR corresponding to different H/O ratios obtained by MPS and HOOI.}
	\label{tableCOIL100}
	\centering 
	\begin{tabular}{l l l l |l l l}
Algorithm & CSR & $N_f$& $\epsilon$ & CSR & $N_f$& $\epsilon$\\
		\hline
		\underline{$r=50\%$}&&&&\underline{$r=85\%$}&&\\
		HOOI & $98.87\pm 0.19$ & $198$ & $0.80$&$94.13\pm 0.42$&$112$&$0.75$\\	
		MPS & $\bf{99.19\pm 0.19}$ & $120$ & $0.80$&$\bf{95.37\pm 0.31}$&$18$&$0.65$\\	 \underline{$r=90\%$}&&&&\underline{$r=95\%$}&&\\
	HOOI & $87.22\pm 0.56$ & $112$&$0.75$&$77.76\pm 0.90$&$112$&$0.75$\\	
		MPS& $\bf{89.38\pm 0.40}$ & $59\pm 5$&$0.75$&${\bf 83.17\pm 1.07}$&$18$&$0.65$\\			
		\hline
	\end{tabular}
\end{table*}

We also perform experiment on the EYFB dataset which contains 16128 grayscale images with 28 human subjects, under 9 poses, where for each pose there is 64 illumination conditions. Similar to \cite{LI14}, to improve computational time each image was cropped to keep only the center area containing the face, then resized to 73 x 55. The training and test datasets are not selected randomly but partitioned according to poses. More precisely, the training and test datasets are selected to contain poses 0, 2, 4, 6 and 8 and 1, 3, 5, and 7, respectively. For a single subject the training tensor has size $5\times73\times55\times64$ and $4\times73\times55\times64$ is the size of the test tensor. Hence for all 28 subjects we have fourth-order tensors of sizes $140\times73\times55\times64$ and $112\times73\times55\times64$ for the training and test datasets, respectively.
\begin{table*}[!htbp]
	\caption{EYFB benchmark with reduced core tensors being used for classification, core sizes of MPS and HOOI are $Q\times\De\times\De$ and $Q\times\De\times\De\times 1$, where $\De\in(10,\ldots,14)$, respectively.}
	\label{tablecEYFB}
	\centering 
	\begin{tabular}{l l l l l}
		\multirow{1}{*}{Algorithm} & CSR ($\epsilon = 0.9$)&  CSR ($\epsilon = 0.85$)&
		CSR ($\epsilon = 0.80$)
		& CSR ($\epsilon = 0.75$) \\				
		\hline
		{\bf KNN-1}&&&&\\		
		HOOI & $90.71\pm 1.49$ & $90.89\pm 1.60$ & $91.61\pm 1.26$ & $88.57\pm 0.80$\\
		MPS & ${\bf 94.29\pm 0.49}$ & ${\bf94.29\pm 0.49}$ & ${\bf 94.29\pm 0.49}$ & ${\bf 94.29\pm 0.49}$\\	
		{\bf LDA}&&&&\\		
		HOOI & $96.07\pm 0.80$ & $95.89\pm 0.49$ & $96.07\pm 0.49$ & $96.07\pm 0.49$\\
		MPS & ${\bf 97.32\pm 0.89}$ & ${\bf97.32\pm 0.89}$ & ${\bf 97.32\pm 0.89}$ & ${\bf 97.32\pm 0.89}$\\	
		\hline
	\end{tabular}
\end{table*}

\begin{table*}[!ht]
	\caption{BCI Jiaotong benchmark with reduced core tensors being used for classification, core sizes of MPS and HOOI are $Q\times 12\times\De$ and $Q\times 12\times\De\times 1$, where $\De\in(8,\ldots,14)$, respectively.}
	\label{tablecBCI}
	\centering 
	\begin{tabular}{l l l l l}
		\multirow{1}{*}{Algorithm} & CSR ($\epsilon = 0.9$)&  CSR ($\epsilon = 0.85$)&
		CSR ($\epsilon = 0.80$)
		& CSR ($\epsilon = 0.75$) \\				
		\hline
		{\bf Subject 1}& & & &\\		
		HOOI & $84.39\pm 1.12$ & $83.37\pm 0.99$ & $82.04\pm 1.05$ & $84.80\pm 2.21$\\
		MPS & ${\bf 87.24\pm 1.20}$ & ${\bf 87.55\pm 1.48}$ & ${\bf 87.24\pm 1.39}$ & ${\bf 87.65\pm 1.58}$\\	
		{\bf Subject 2}& & & &\\		
		HOOI & $83.16\pm 1.74$ & $82.35\pm 1.92$ & $82.55\pm 1.93$ & $79.39\pm 1.62$\\
		MPS & ${\bf 90.10\pm 1.12}$ & ${\bf 90.10\pm 1.12}$ & ${\bf 90.00\pm 1.09}$ & ${\bf 91.02\pm 0.70}$\\		
		\hline
	\end{tabular}
\end{table*}
We apply the MPS and HOOI to extract the core tensors before inputting them into the classifiers. However, in this experiments we realize that the size of each core tensor remains very large even with small threshold used, e.g., for $\epsilon=0.75$, the core size of each sample obtained by MPS and HOOI are $18\times 201 = 3618$ and $14\times 15\times 13 = 2730$, respectively which is not useful for directly classifying. Therefore, we need to further reduce the sizes of core tensors before feeding them to classifiers for a better performance. In our experiment, we simply apply a further truncation to each core tensor by keeping the first few dimensions of each mode of the tensor. Intuitively, this can be done as we have already known that the space of each mode is orthogonal and ordered in such a way that the first dimension corresponds to the largest singular value, the second one corresponds to the second largest singular value and so on. Therefore, we can independently truncate the dimension of each mode to a reasonably small value (which can be determined empirically) without changing significantly the meaning of the core tensor. It then gives rise to a core tensor of smaller size that can be used directly for classification. More specifically, suppose that the core tensors obtained by MPS and HOOI have sizes $Q\times\De_1\times\De_2$ and $Q\times\De_1\times\De_2\times\De_3$, where $Q$ is the number $K$ ($L$) of training (test) samples, respectively. The core tensors are then truncated to be $Q\times\tilde{\De}_1\times\tilde{\De}_2$ and $Q\times\tilde{\De}_1\times\tilde{\De}_2\times\tilde{\De}_3$, respectively such that $\tilde{\De}_l<\De_l$ ($l=1,2,3$). Note that each $\tilde{\De}_l$ is chosen to be the same for both training and test core tensors.
We show the classification results for different threshold values $\epsilon$ in Table.~\ref{tablecEYFB} using two different classifiers, i.e. KNN-1 and LDA. In this result,  the core tensors obtained by MPS and HOOI are reduced to have sizes of $Q\times\tilde{\De}_1\times\tilde{\De}_2$ and $Q\times\tilde{\De}_1\times\tilde{\De}_2\times\tilde{\De}_3$, respectively such that $\tilde{\De}_1=\tilde{\De}_2=\De\in(10,11,12,13,14)$ and $\tilde{\De}_3=1$. As a result, the reduced core tensors obtained by both methods have the same size for classification. Note that each value of CSR in Table.~\ref{tablecEYFB} is computed by taking the average of the ones obtained from classifying different reduced core tensors due to different $\De$. We can see that the MPS gives rise to better results for all threshold values using different classifiers. More importantly, MPS with smallest $\epsilon$ can also produce CSR as well as largest $\epsilon$. The LDA classifier gives rise to the best result, i.e. ${\bf 97.32\pm 0.89}$.

Lastly, we study the BCIJ dataset which consists of single trial recognition for BCI electroencephalogram (EEG) data involving left/right motor imagery (MI) movements. The original experiment had five subjects and the paradigm required subjects to control a cursor by imagining the movements of their right or left hand for 2 seconds with a 4 second break between trials. Subjects were required to sit and relax on a chair, looking at a computer monitor approximately 1m from the subject at eye level. For each subject, data was collected over two sessions with a 15 minute break in between. The first session contained 60 trials (30 trials for left, 30 trials for right) and were used for training. The second session consisted of 140 trials (70 trials for left, 70 trials for right). The EEG signals were sampled at 500Hz and preprocessed with a filter at 8-30Hz, hence for each subject the data consisted of a multidimensional tensor $channel\times time\times trial$. Prior to simulation we preprocessed the data by transforming the tensor into the time-frequency domain using complex Mortlet wavelets with bandwidth parameter $f_b=6$Hz (CMOR6-1) to make classification easier \cite{ZHA07, PHA11}. The wavelet center frequency $f_c=1$Hz is chosen. Hence, the size of the concatenated tensors are $62\ channels\times 23\ frequency\ bins\times 50\ time\ frames\times Q$.

We perform the experiment for subject 1 and 2 of the dataset. Similar to the case of the EYFB dataset, after applying the feature extraction methods, the core tensors still have high dimension, so we need to further reduce their sizes before using them for classification. For instance, the reduced core sizes of MPS and HOOI are  chosen to be $Q\times 12\times\De$ and $Q\times 12\times\De\times 1$, where $\De\in(8,\ldots,14)$, respectively. For classification, we use LDA and the classification results are shown in Table.~\ref{tablecBCI} for different threshold values. We can see that MPS always performs better than HOOI.

\section{Conclusion}\label{Conclusions}
In this paper, we propose MPS as an alternative to TD-based algorithms for feature extraction applied to tensor classification problem. Compared to HOOI, MPS has been shown to have some advantages such as computational savings due to successive SVDs are employed and no recursive optimization needed for acquiring common factors and core tensors. In addition, using extracted features given by core tensors for classifiers is capable of leading to better classification success rate even though a same number of features is used from both MPS and HOOI. We have validated our method by applying it to classify a few multidimensional datasets, such as visual data (COIL-100 and EYFB) and EEG signals where training and test data represented by fourth-order tensors. Benchmark results show that MPS gives better classification rates than HOOI in most cases.

For the future outlook, we plan to further improve MPS for classifying very big multilinear datasets. We also plan to extend this promising tool for many other problems such as multilinear data compression and completion.

\bibliographystyle{IEEEtran}
\bibliography{ref}
\vfill
\end{document}